\pdfoutput=1

\documentclass[11pt]{article}

\usepackage[]{EACL2023}

\usepackage{times}
\usepackage{latexsym}

\usepackage[T1]{fontenc}

\usepackage[utf8]{inputenc}

\usepackage{microtype}

\usepackage{inconsolata}
\usepackage{booktabs}
\usepackage{multirow}
\usepackage{graphicx}
\usepackage{tablefootnote}
\usepackage{hyperref}
\usepackage{caption}
\usepackage{subcaption}
\usepackage{float}
\usepackage[ruled, linesnumbered]{algorithm2e}
\SetStartEndCondition{ }{}{}%
\SetKwProg{Fn}{def}{\string:}{}
\SetKwFunction{Range}{range}
\SetKwFunction{Any}{any}
\SetKw{KwTo}{in}
\SetKwFor{For}{for}{\string:}{}%
\SetKwIF{If}{ElseIf}{Else}{if}{:}{elif}{else:}{}%
\SetKwFor{While}{while}{:}{}%

\SetKw{KwIs}{is}
\SetKw{KwNot}{not}
\SetKwFunction{build}{build\_tree}%
\SetKwFunction{child}{get\_child}%
\SetKwFunction{generate}{generate\_candidates}%
\SetKwFunction{rank}{tree\_rank}%
\SetKwFunction{select}{select}%
\SetKwFunction{syntax}{syntax\_select}%
\SetKwFunction{bfs}{level\_order}%
\SetKwFunction{prune}{prune}%
\SetKwData{Null}{NULL}
\SetKwData{Node}{Node}\SetKwData{Root}{root}\SetKwData{Up}{up}
\title{More Robust Schema-Guided Dialogue State Tracking via Tree-Based Paraphrase Ranking}

\author{Alexandru Coca$^{\dagger}$, Bo-Hsiang Tseng$^{\ddagger}$, Weizhe Lin$^{\dagger}$, Bill Byrne$^{\dagger}$ \\
        ${}^\dagger$Department of Engineering, University of Cambridge, United Kingdom \\
        ${}^\ddagger$Apple \\
        \texttt{\{ac2123, wl356, wjb31\}@cam.ac.uk} \\
        \texttt{bohsiang\_tseng@apple.com} }

\begin{document}
\maketitle
\begin{abstract}
The schema-guided paradigm overcomes scalability issues inherent in building task-oriented dialogue (TOD) agents with static ontologies. Instead of operating on dialogue context alone, agents have access to hierarchical schemas containing task-relevant natural language descriptions. Fine-tuned language models excel at schema-guided dialogue state tracking (DST) but are sensitive to the writing style of the schemas. We explore methods for improving the robustness of DST models. We propose a framework\footnote{Code will be released here: \url{https://bit.ly/3WYB7Fl}} for generating synthetic schemas which uses tree-based ranking to jointly optimise lexical diversity and semantic faithfulness. The generalisation of strong baselines is improved when augmenting their training data with prompts generated by our framework, as demonstrated by marked improvements in average joint goal accuracy (JGA) and schema sensitivity (SS) on the SGD-X benchmark.
\end{abstract}

\section{Introduction}
\label{sec:intro}
DST is concerned with tracking user goals in task-oriented conversations. The goals are represented as key-value pair sequences, with the keys known as \textit{slots} (e.g. hotel name). Pre-trained language models (PLMs) \citep{DBLP:conf/naacl/DevlinCLT19, DBLP:journals/jmlr/RaffelSRLNMZLL20} have helped shift focus from systems that can only track slots drawn from a database or \textit{domain ontology} \citep{DBLP:conf/sigdial/HendersonTY14} to models that do not require re-training to parse goals in new domains. The Schema-Guided Dialogue (SGD) dataset \citep{DBLP:conf/aaai/RastogiZSGK20} facilitates this shift with a large-scale set of conversations grounded in $45$ \textit{service APIs} or \textit{schemas} that describe the domains, slots and user intents that annotate the conversations (Appendix \ref{appendix:sgd-sgd-x}). Test set dialogues are grounded in $6$ schemas \textit{seen} during training and $15$ \textit{unseen} ones. 

Neural models perform impressively on the difficult schema-guided DST task \cite{DBLP:conf/aaai/RastogiZSGK20}, but \citet{DBLP:conf/aaai/0001GR0ZW22} show that the uniformity of the descriptive language of the schemas facilitates this. They create the SGD-X benchmark to evaluate robust zero-shot generalisation of DST models. This is achieved by grounding the SGD test set conversations in five \textit{schema variants} increasingly dissimilar to the SGD schemata\footnote{Variants are ordered according to their lexical similarity to the SGD schemas. The \texttt{v1} variant is the most similar whereas \texttt{v5} is the most dissimilar. See Appendix \ref{appendix:sgd-sgd-x} for details and examples and the schemata here: \href{https://github.com/google-research-datasets/dstc8-schema-guided-dialogue/tree/master/sgd_x/data/v1}{https://bit.ly/3Ev0KrV}.}. To perform well, a DST model should correctly track the state of a dialogue when conditioned, in turn, on prompts constructed from the five variants. 

We show how to improve DST robustness by introducing controlled variability in the data. We contribute to robust DST research by (1) a flexible framework for generating and ranking diverse outputs of a paraphrase model based on a tree-clustering algorithm designed to control lexical diversity and semantic similarity; (2) combine  state-of-the-art paraphrase models and language generation metrics to generate increasingly diverse schemata paraphrases; (3) show that augmenting the training dataset with these schemata improves the robustness and generalisation performance of strong DST baselines.

\section{Related Work}
\label{sec:background}

Input variety, data scarcity and domain shifts affect the robustness of DST models. \citet{DBLP:conf/acl/LiuTWWLNLPH20} investigate the former. They employ word-level data augmentation (DA) \cite{DBLP:conf/emnlp/WeiZ19}, turn paraphrasing and speech disfluency modeling to approximate their field performance. Turn and dialogue generation are effective in low-resource settings \citep{DBLP:conf/acl/CampagnaFML20, DBLP:conf/coling/HouLCL18} but are very difficult to scale to new domains and are not effective in the high-resource setting we consider \cite{DBLP:conf/acl/CampagnaFML20, DBLP:conf/emnlp/Mohapatra0CJ21}. This also applies to word- and sentence-level methods \cite{DBLP:conf/ialp/QuanX19, DBLP:conf/paclic/LouvanM20}. \citet{DBLP:conf/aaai/0001GR0ZW22} find word-order changes and deletions to be ineffective in the high-resource, schema-guided setting we consider. 

Schema-guided DST tackles both data scarcity and novel domains by using API definitions to prompt PLMs \cite{DBLP:journals/corr/abs-2201-08904}. Yet \citet{DBLP:conf/aaai/0001GR0ZW22} demonstrate the lack of robustness of schema-guided DST models to prompt styles and vocabulary, creating a new research direction. They show that augmenting the training data with synthetic prompts obtained via backtranslation significantly improves models' ability to track states under meaning-preserving prompt transformations. Backtranslation is also applied to improve DST robustness to linguistic variation inherent in user communication \cite{DBLP:journals/corr/abs-1912-09297, DBLP:journals/corr/abs-1911-05153}, which is orthogonal to the prompt style and vocabulary robustness setting we consider. Reinforcement learning has also been applied \cite{DBLP:conf/aaai/YinSJCL20}, but works only in the very constrained single-domain, ontology-driven setting. Other TOD-relevant DA approaches apply to policy learning \cite{DBLP:journals/tacl/GrittaLI21} and response-generation \cite{DBLP:conf/acl/GaoZOY20, DBLP:conf/aaai/ZhangOY20}.

Addressing the dearth of augmentation methods designed to ensure prompt robustness of schema-guided DST models, we propose to generate schemas by ranking large paraphrase candidate lists with learned metrics in a tree ranking scheme. 

\begin{figure*}
	\centering
	\includegraphics[width=\textwidth]{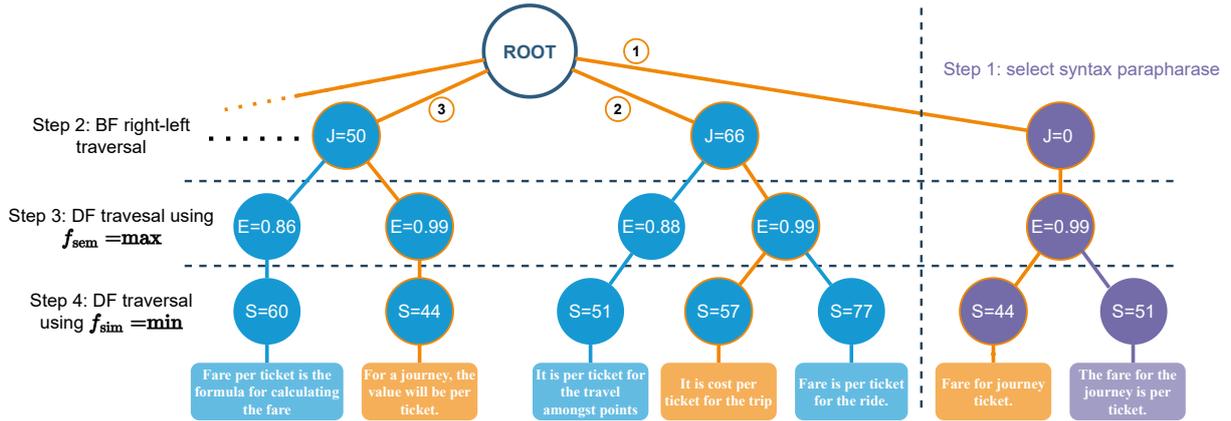}
	\caption{Ranking paraphrases of \textit{Fare per ticket for journey} using a tree. Top level node split is by Jaccard distance ($J$), middle nodes split by entailment score ($E$) and leave nodes store string similarities ($S$). Here $f_{\mbox{dec}} = [\mbox{None}, \max, \min]$. By using $J$ we guarantee that candidates with $J=0$ are syntactic paraphrases if $S$ is constrained. Orange leaves show top ranked candidates. Numbers on paths show ranking order.
}
	\label{fig:main-body-rankigschematic}
\end{figure*}
\section{Tree-Based Paraphrase Ranking}
\label{sec:method}

\textbf{Tree construction} A large pool of schema candidates is created by generating paraphrases given grids of generation parameters (eg temperature, number of beams). The set is filtered to address generation failures (eg toxic and hallucinated words). We optionally filter candidates with an entailment model to increase semantic faithfulness \citep{DBLP:conf/acl/NarayanSZM00L22} (see Appendix \ref{appendix:cand-gen}).

The tree constructor (Algorithm \ref{algo:build}) takes as input an object (\texttt{Node}) that stores a metric value, \texttt{val}, and the candidate paraphrases which are split at that node, \texttt{sents}. A list of metrics to be computed between each candidate and the input is provided by the user. This enables our framework to build arbitrary-depth trees with custom user metrics. Each unique list of metric values describing the distance between the input and a candidate generates a path in the tree (lines 5-13). The $n$-ary tree constructed in this way has the property that level-order traversal of the first level can yield diverse candidates with respect to the metric it encodes. In practice, the metrics measure lexical and semantic distances between their inputs.

\begin{algorithm}
\SetNlSty{}{}{:}
\SetAlgoNlRelativeSize{-1}
\AlgoDontDisplayBlockMarkers\SetAlgoNoEnd\SetAlgoNoLine%
\caption{Tree building}
\label{algo:build}
\Fn{\build{\texttt{root}: Node, \texttt{inp}: str, \texttt{cands}: list[str], \texttt{metrics}: list[Callable]}}{
\KwData{\texttt{root}, \texttt{inp} input , \texttt{cands} input paraphrases, \texttt{metrics} objects to eval.  dist. between input \& cand.}
\KwResult{tree splitting \texttt{cands} according to \texttt{metrics}}

\texttt{curr} $\gets$ \texttt{root} \;
\For{\texttt{c} \KwTo \texttt{cands}}{
    \texttt{curr} $\gets$ \texttt{root} \;
    \For{\texttt{m} \KwTo \texttt{metrics}}{
        \texttt{m\_val} = \texttt{m(\texttt{inp}, \texttt{c})} \;
        \texttt{next} $\gets$ \child(\texttt{curr.children}, \texttt{m\_val}) \;
        \eIf{\texttt{next} \KwIs \Null }{ 
                \texttt{next} $\gets$ \Node(val=\texttt{m\_val}, sents=\texttt{[\texttt{c}]}) \;
                \texttt{curr.children.add(next)}
        }{
        \texttt{next.sents.add(c)} \;
        \texttt{curr} $\gets$ \texttt{next}
    }
    }
}

\Return \texttt{root}
}
\end{algorithm}

\textbf{Ranking} Our ranker input is the tree and a list of decision functions, with elements corresponding to each level in the tree, \texttt{f\_dec}. Without loss of generality, we assume that the first level encodes a metric with respect to which the user wishes to maximise diversity (eg lexical distance). As shown in Figure \ref{fig:main-body-rankigschematic}, our algorithm traverses breadth-first the level for which diversity is to be maximised. Each subtree returned in the traversal is traversed depth-first, guided by the decision functions. For example, in Figure \ref{fig:main-body-rankigschematic} we show that the node $B=0.77$ is selected by applying the $\max$ decision function to the children of $J=66$, and that applying $\min$ to the children of $B=0.77$ selects the leaf $S=77$. See Algorithm \ref{algo:ranking} (Appendix \ref{appendix:ranking-framework}) for details.

\section{Experiments}
\label{sec:experiments}
\subsection{Schema generation}
Our paraphrase model is PegasusParaphrase\footnote{Available at \url{https://bit.ly/3vgY7EZ}.}, a fine-tuned Pegasus model \citep{DBLP:conf/icml/ZhangZSL20}. Using $10$ settings for the number of beams and temperature, we generate $500$ candidates for each input. For efficiency purposes, these are filtered heuristically (Appendix \ref{appendix:ranking-framework}). We construct a depth $d=3$ tree which splits the candidates by Jaccard distance $J$, entailment $E$, and string similarity $S$. That is, the input to our tree constructor (Algorithm \ref{algo:generate}) is $\mbox{metrics} = [J,E, S]$. Here entailment is computed using BART \citep{DBLP:conf/acl/LewisLGGMLSZ20} as described in Appendix \ref{appendix:ranking-framework}. We prune nodes with $J > 0.75$ to limit hallucination.

We select $k=5$ lexically-diverse paraphrases that maximise entailment given the constraint that the returned candidates should be lexically diverse. First, we select a syntactic paraphrase by traversing the subtree rooted at $J=0$ and minimising $S$. The remainder of the candidates are selected by constraining the bread-first traversal of the first level, which encodes lexical distance, to return the nodes sorted from high to low. This procedure is depicted schematically in Figure \ref{fig:main-body-rankigschematic}. We sort the ranked candidate lists for each description according to the Jaccard distance between them and the SGD descriptions. Hence we obtain $k=5$ synthetic schema variants, with $v1$ being the most  similar to the SGD schema and $v5$ the most dissimilar. We refer to this scheme as \texttt{Pegasus + BART}. 

\subsection{State tracking data augmentation}
\subsubsection{Baseline models}

\textbf{D3ST} The \underline{D}escription-\underline{D}riven \underline{D}ialogue Modelling (D3ST) model \citep{DBLP:journals/corr/abs-2201-08904} is a state-of-the-art DST model that performs intent tracking, requested slots prediction, and state tracking in a single pass. See Appendix \ref{appendix:state-tracking} for a visual representation of inputs and targets. We process the data and train the model as described in \citet{DBLP:journals/corr/abs-2201-08904} and Appendix \ref{appendix:state-tracking}, selecting models that maximise the development set JGA.

\textbf{T5DST} We follow \citet{DBLP:conf/aaai/0001GR0ZW22} to implement a simplified T5DST \citep{DBLP:conf/emnlp/00010O21}. It predicts the value of each slot iteratively, requiring a number of decoding passes equal to the number of slots in an API to predict the dialogue state given a dialogue history. Training and inference with this model is very expensive and we train the models with a fixed computational budget of $20,000$ gradient steps\footnote{This is the number of steps required for maximising the development set JGA, for all three runs.} for the baseline and $40,000$ steps for all augmented data experiments. See Appendix \ref{appendix:state-tracking} for prompt structure and implementation details.

\subsubsection{Evaluation}
On SGD, the JGA ($\mbox{JGA}_{orig}$) is computed for the $4,201$ test set dialogues. $77\%$ of these have a turn span where the agent calls an API unseen in training. Only $6$ out of $21$ schemata are seen in training. 

Evaluation on SGD-X proceeds as follows. First, the SGD descriptions in the prompt are replaced, in turn, with descriptions taken from the five SGD-X variants. The DST model then predicts the state of a given dialogue $5$ times, conditioned on prompts that are increasingly dissimilar to the SGD test set. Hence, the $\mbox{JGA}_{v_{1-5}}$ figures reported are averages over approximately $21, 000$ conversations. For all experiments except \textit{oracle} (see Section \ref{sec:experiment-desc}), \textit{none} of the test time prompts are seen during training: the \textit{seen} superscript in the metric names reported in Section \ref{sec:results} identifies conversations where the \textit{SGD test set prompt is seen} during training. Therefore, it quantifies whether the model can robustly identify slots seen in training by interpreting the meaning of the descriptions rather than relying on linguistic patters in the training schema. Meanwhile \textit{unseen} measures the ability of the model to generalise to new APIs, which may describe new slots and domains, notwithstanding the language used by developers to phrase the descriptions. The JGA coefficient of variation (ie \textit{schema sensitivity}, $\mbox{SS}_{JGA}$) as the prompt changes measures the sensitivity of a model to the prompt \cite{DBLP:journals/corr/abs-2110-06800}.
\begin{table}[H]
\centering
\resizebox{\columnwidth}{!}{%
\begin{tabular}{ccllllc}
\toprule
Metric &
  Ranking &
  \multicolumn{1}{c}{v1} &
  \multicolumn{1}{c}{v2} &
  \multicolumn{1}{c}{v3} &
  \multicolumn{1}{c}{v4} &
  v5 \\ \hline
\multirow{4}{*}{Jaccard Dist} &
  Pegasus + BART &
  17.1 &
  63.0 &
  69.5 &
  72.0 &
  \multicolumn{1}{l}{76.6} \\
                            & Backtranslation & 18.2                  & 29.9 & 43.9 & \multicolumn{1}{c}{-} & -                        \\
                            & EDA             & 3.9                   & 4.1  & 6.1  & 16.0                  & \multicolumn{1}{l}{32.9} \\
                            & SGD-X           & 55.6                  & 65.6 & 71.2 & 78.1                  & \multicolumn{1}{l}{85.7} \\ \hline
\multirow{4}{*}{Entailment} & Pegasus + BART  & 99.0                  & 96.7 & 94.9 & 94.6                  & 94.4                     \\
                            & Backtranslation & 97.5                  & 96.5 & 95.9 & \multicolumn{1}{c}{-} & -                        \\
                            & EDA             & 99.1                  & 98.5 & 96.6 & 93.2                  & \multicolumn{1}{l}{86.4} \\
                            & SGD-X           & 89.7                  & 88.0 & 88.4 & 86.8                  & 87.5                     \\ \hline
\multirow{4}{*}{BLEU}       & Pegasus + BART  & 13.4                  & 12.5 & 12.5 & 13.2                  & 12.7                     \\
                            & Backtranslation & 36.4                  & 26.0 & 18.9 & \multicolumn{1}{c}{-} & -                        \\
                            & EDA             & 72.0                  & 63.3 & 47.2 & 42.3                  & \multicolumn{1}{l}{44.2} \\
                            & SGD-X           & 20.4                  & 15.3 & 10.8 & 8.3                   & 5.2                      \\ \hline
\multirow{4}{*}{self-BLEU} &
  Pegasus + BART &
  \multicolumn{1}{c}{-} &
  12.0 &
  11.4 &
  11.0 &
  10.9 \\
 &
  Backtranslation &
  \multicolumn{1}{c}{-} &
  49.3 &
  41.7 &
  \multicolumn{1}{c}{-} &
  - \\
                            & EDA             & \multicolumn{1}{c}{-} & 87.0 & 68.8 & 58.0                  & \multicolumn{1}{l}{53.1} \\
                            & SGD-X           & \multicolumn{1}{c}{-} & 13.5 & 11.2 & 9.9                   & 8.6                      \\ \bottomrule
\end{tabular}
}
\caption{Automatic synthetic schema evaluation. $J$ is multiplied by $100$ for readability.}
\label{tab:schema-generations-metric}
\vspace{-0.35cm}
\end{table}
\subsubsection{Experimental setup}
\label{sec:experiment-desc}
We show our approach is effective by augmenting the DST training data with synthetic prompts composed from our generated schemata. To study the effect of controlling prompt diversity augmented datasets are two (2x) to six times (6x) the SGD size. For 2x, augmented data contains prompts constructed from the \texttt{v1} synthetic schema, whereas for 6x we use all five generated schemas\footnote{Ordering is from most (\textit{v1}) to least (\textit{v5}) similar to SGD.}.  

\textbf{Baselines} We create three synthetic schema by \textbf{backtranslation}. Our pivot languages are Korean, Japanese and Chinese \citep{DBLP:conf/aaai/0001GR0ZW22}. The augmented DST training dataset is four times (4x) larger than SGD. Following \citet{DBLP:conf/eacl/HuangLQP21}, we also consider French and Russian as pivot languages to generate two more synthetic schemas and obtain an augmented dataset six times (6x) larger than SGD. We also compare with \textbf{easy data augmentation} (EDA) \citep{DBLP:conf/emnlp/WeiZ19}, a word-level DA approach based on synonym replacement (SR), random insertion, deletion and substitution. We perform SR with probability $0.25$ and the other operations with equal probability of $0.05$. Just like for backtranslation, we generate $3$ or $5$ synthetic schemas with this method via the public API. Augmentation with the \textbf{SGD-X} human schemata paraphrases is considered an \textit{oracle} because these models see the SGD-X schemata at training time.

\begin{table*}[h]
\centering
\small
\vspace{-0.3cm}
\begin{tabular}{cccccccc}
\toprule
Model & \# & 
  Generation method - Dataset size &
  $\mbox{JGA}_{orig} \uparrow $ &
  $\mbox{JGA}_{v_{1-5}}$ &
  $\mbox{JGA}_{v_{1-5}}^{seen}$ &
  $\mbox{JGA}_{v_{1-5}}^{unseen}$ &
  $\mbox{SS}_{JGA} \downarrow $ 
  \\ \hline
                  & 1   & None - 1x                                                          & 69.8          & 56.5          & 73.6          & 50.8          & 70.1          \\ \cline{2-8} 
                & 2     & Pegasus + BART - 2x                                             & \textbf{72.8} & 61.3          & 80.9          & 54.8          & 56.4         \\
                 & 3    & Pegasus + BART - 4x                                             & 72.6          & 62.5          & 81.7          & 56.1          & 51.0        \\
                & 4     & Pegasus + BART  - 6x                                             & 71.2          & \textbf{63.9} & \textbf{85.9} & \textbf{56.6} & \textbf{46.5}  \\ \cline{2-8} 
& 5 &
  {\color[HTML]{000000} EDA - 4x  \citep{DBLP:conf/emnlp/WeiZ19}} &
  {\color[HTML]{000000} 71.0} &
  {\color[HTML]{000000} 59.0} &
  {\color[HTML]{000000} 78.5} &
  {\color[HTML]{000000} 52.5} &
  {\color[HTML]{000000} 63.0}\\
                   & 6  & {\color[HTML]{000000} EDA - 6x  \citep{DBLP:conf/emnlp/WeiZ19}} & 71.4          & 62.3          & 83.3          & 55.3          & 53.2        \\
                  & 7   & Backtranslation - 4x \citep{DBLP:journals/corr/abs-2110-06800}  & 72.1          & 62.2          & 84.0          & 54.9          & 53.1       \\
                   & 8  & Backtranslation - 6x \citep{DBLP:conf/eacl/HuangLQP21}                                          & 71.5          & 61.0          & 82.5          & 53.8          & 54.4         \\ \cline{2-8} 
\multirow{-9}{*}{D3ST} & 9 & 
  \textit{SGD-X} - 6x \citep{DBLP:journals/corr/abs-2110-06800} \textit{(Oracle)} &
  {{73.8}} &
  {{69.7}} &
  {{92.5}} &
  {{62.1}} &
  {{27.9} }\\ \hline
\multicolumn{1}{l}{} & 10 & None                                                          & 70.0          & 50.4          & 58.5          & 47.7          & 87.0          \\ \cline{2-8} 
\multicolumn{1}{l}{}& 11 & Pegasus B + BART - 4x                                           & 71.3          & \textbf{55.1} & 71.2          & \textbf{49.7} & \textbf{70.1}  \\
\multicolumn{1}{l}{} & 12 & Pegasus + BART - 6x                                            & 68.7          & 52.5          & \textbf{71.6} & 46.5          & 77.6         \\ \cline{2-8} 
\multicolumn{1}{l}{}& 13 & EDA - 6x \citep{DBLP:conf/emnlp/WeiZ19}                         & 72.2          & 51.1          & 55.6          & 49.6          & 84.1         \\
\multicolumn{1}{l}{} & 14 &
  Backtranslation - 4x \citep{DBLP:journals/corr/abs-2110-06800} &
  \textbf{72.8} &
  53.9 &
  67.0 &
  49.6 &
  76.4 \\ \cline{2-8} 
\multicolumn{1}{l}{\multirow{-6}{*}{T5DST}} & 15 &
  \textit{SGD-X} - 6x \citep{DBLP:journals/corr/abs-2110-06800} \textit{(Oracle)} &
  {{74.2}} &
  {{67.2}} &
  {{91.8}} &
  {{59.0}} &
  {{36.6}} \\ \bottomrule
\end{tabular}
\vspace{-0.1cm}
\caption{SGD and SGD-X dialogue state tracking performance when training with augmented data. Best performance (excluding the oracle setup) is in \textbf{bold}. Dataset size is the number of times the augmented dataset is larger than SGD.}
\label{tab:dst-results}
\vspace{-0.4cm}
\end{table*}
\vspace{-0.1cm}
\section{Results and Discussion}
\label{sec:results}
\subsection{Synthetic schema generation}

Our ranking method generates increasingly lexically diverse schemata as shown by the increase in Jaccard distance across schema variants (Table \ref{tab:schema-generations-metric}). This aspect is much more difficult to achieve with EDA without significantly affecting semantics. Furthermore, self-BLEU \citep{DBLP:conf/sigir/ZhuLZGZWY18} scores indicate EDA is the least effective in ensuring candidate diversity compared to other approaches. The BLEU difference between the SGD-X variants \texttt{v1} and \texttt{v5} is $15.2$ but smaller ($0.66$) for our approach. Hence, the \texttt{PEGASUS + BART} copies $n$-grams from the input and includes additional information. This information is not always meaning-preserving: \textit{City where the event is happening} is paraphrased as \textit{The bustling city where the event is taking place} ($v5$) but \textit{End date for the reservation or to find the house} is paraphrased as \textit{End date for hotel reservation to allow time for a replacement both at the struck and in the run up to the event} ($v5$). The self-BLEU  of the SGD-X schemas decreases faster compared to the automatically generated paraphrases, suggesting that Jaccard distance increases partly due to hallucination.

Entailment scores show that backtranslation is effective in preserving semantics. For EDA, the semantic similarity drops significantly as more candidates are generated since more dissimilar schemas are generated with more edit operations which are likely to affect meaning. The entailment scores for the SGD-X paraphrases are also lower since they do not always perfectly semantically overlap with the input by construction \cite{DBLP:conf/aaai/0001GR0ZW22} and because of entailment model errors.

\subsection{Dialogue state tracking}

\textbf{D3ST} Both the robustness and robust generalisation are improved by augmentation with our synthetic schemas, as demonstrated by maximum $\mbox{JGA}^{seen}_{v_{1-5}}$ ($12.35\%$) and $\mbox{JGA}^{unseen}_{v_{1-5}}$ ($5.85\%$)  increases and $23.6\%$ drop in $\mbox{SS}_{JGA}$ (rows 1\&4, Table \ref{tab:dst-results}). The most benefit is obtained by training with syntactically diverse prompts (\texttt{Pegasus + BART 2x}). Adding more diverse data (rows 3\&4) improves DST performance. Part of this improvement may arise because paraphrasing leaves out domain-dependent information: \textit{Average review rating of the doctor} is paraphrased as \textit{The rating is average, so it's not perfect}, so the model can learn to identify ratings more generally\footnote{It appears in $4$ unseen services in the test set.}. Moreover, inputs are noisy due to hallucination, so the models trained with our augmentation are less likely to overfit to the linguistic patterns of the training schemas. BLEU scores indicate high lexical overlap between EDA-generated and SGD schemas (Table \ref{tab:schema-generations-metric}). This limits the magnitude of EDA improvement (row 5) and we perform better with less data (rows 3\&6, 2\&5). 

Backtranslation is comparable with our method given the same data quantity (rows 3\&7). When we also backtranslate via French and Russian \citep{DBLP:conf/eacl/HuangLQP21} the data diversity does not significantly increase (Table \ref{tab:backtr-4x-6x}, Appendix \ref{appendix:additional-results}). This negatively impacts the DST performance, while our method improves it (rows 4\&8). We can control the schema generation process to match SGD backtranslation performance (Appendix \ref{appdx:QCPG-BLEURT-RESULTS}).

\textbf{T5DST}\footnote{\citet{DBLP:journals/corr/abs-2110-06800} report $72.6\%$ JGA on SGD and $64.0\%$ SGD-X but we could reproduce only $69.98\%$ and $50.42\%$.} We outperform EDA (rows 12\&13) but not backtranslation (row 14). This may be due to (1) the larger computational budget needed to maximise T5DST performance\footnote{Each training example is seen only once.} and (2) T5DST's sensitivity to noisy descriptions owing to its prompt format (Appendix \ref{appendix:state-tracking}). We control hallucination by pruning candidates with $J > 0.5$ and entailment smaller than $0.58$ and maximise $J$ while minimising $S$ to produce an augmented dataset 4x larger than SGD (\texttt{Pegasus B + BART 4x}). Limiting lexical diversity improves entailment compared to \texttt{Pegasus + BART 6x} (Table \ref{tab:ranking-algos-comparison-slots}, Appendix \ref{appendix:additional-results}), and the scheme improves DST robustness compared to the backtranslation baseline (rows 11\&14).

The best augmentation schemes fail to improve robustness and generalisation relative to the human baseline (rows 9\&15). This is due to the intrinsic challenge of generating diverse yet semantically faithful paraphrases but also due to the fact that humans use common sense and schema information when paraphrasing, so the SGD-X paraphrases are not strictly semantically equivalent. However, the proposed automatic process of paraphrase generation enhances DST, yielding non-trivial improvements in model robustness, while being less costly and more scalable compared to gathering human-written schemata paraphrases.

\vspace{-0.1cm}
\section{Conclusion and Future Work}
\vspace{-0.1cm}
We presented a simple tree-based ranking algorithm for optimising lexical diversity and semantic faithfulness during schema generation. The synthetic schemas improve both the DST models' robustness to  schemata writing style and their generalisation. Our framework will allow researchers working on paraphrase generation and semantic faithfulness to measure the generalisation of their models in a way that may be difficult to capture by existing benchmarks: it can generate schemata paraphrases and train SOTA dialogue state trackers which were shown to benefit from augmentation with high quality, crowdsourced paraphrases.

\section*{Limitations}
The optimality of our ranking method depends on the ability of the underlying paraphrase model to generate a search space that contains paraphrases which are lexically and syntactically diverse and preserve the meaning of the input description. This is sometimes challenging with schema inputs which tend to be short (e.g. \textit{name of event}) and contain little information. Our future work will focus on addressing this by contextualising these inputs to enable the paraphrase model to produce a richer space of candidates. Secondly, our method requires that the semantic faithfulness metrics capture semantic similarity well even as the vocabulary of the candidates and their syntax are very diverse. Previous work on abstractive summarisation \cite{DBLP:conf/acl/NarayanSZM00L22, DBLP:conf/acl/MaynezNBM20, kryscinski-etal-2019-neural} finds entailment scores to be best correlated with human judgment of faithfulness. However, the correlations are not perfect so the output of the ranking algorithm is still expected to contain noisy candidates. For slot description paraphrases, this is challenging because different inputs are very closely semantically related and the entailment model may not identify paraphrase model errors that map a slot description (e.g. departure time) to one with related semantics (e.g. arrival time). We intend to address this in future work by developing finetuning schemes for semantic faithfulness metrics.

\section*{Ethics Statement}
Our work is concerned with the use of language generation models to augment training datasets for schema-guided dialogue datasets. The generation phase is unconstrained, so the model may generate candidates that exhibit biases inherited from the C4 \cite{DBLP:journals/jmlr/RaffelSRLNMZLL20} and HugeNews \cite{DBLP:conf/icml/ZhangZSL20} pre-training datasets. In our experiments, we did not observe toxic or harmful outputs, but on one occasion the model did generate the word \textit{apartheid} as part of an incoherent sentence. For this reason, our filtering stack rejects any candidates containing sensitive words. The list of words that parameterize the sensitive words filter is defined by the user.

\section*{Acknowledgements}
Alexandru Coca was supported supported by EPSRC grant EP/R513180/1. He would like to acknowledge Harrison Lee and Raghav Gupta from Google Research for support and guidance with D3ST and T5DST implementation. Weizhe Lin was supported by a Research Studentship funded by Toyota Motor Europe
(RG92562(24020)). We also thank Howard Mei from University of Cambridge for help with editing the final draft. Authors would like to acknowledge the improvement suggestions made by anonymous reviwers and the EACL program committee. 

\bibliography{anthology,custom}

\begin{thebibliography}{32}
\expandafter\ifx\csname natexlab\endcsname\relax\def\natexlab#1{#1}\fi

\bibitem[{Bandel et~al.(2022)Bandel, Aharonov, Shmueli{-}Scheuer, Shnayderman,
  Slonim, and Ein{-}Dor}]{DBLP:conf/acl/BandelASSSE22}
Elron Bandel, Ranit Aharonov, Michal Shmueli{-}Scheuer, Ilya Shnayderman, Noam
  Slonim, and Liat Ein{-}Dor. 2022.
\newblock \href {https://doi.org/10.18653/v1/2022.acl-long.45} {Quality
  controlled paraphrase generation}.
\newblock In \emph{Proceedings of the 60th Annual Meeting of the Association
  for Computational Linguistics (Volume 1: Long Papers), {ACL} 2022, Dublin,
  Ireland, May 22-27, 2022}, pages 596--609. Association for Computational
  Linguistics.

\bibitem[{Campagna et~al.(2020)Campagna, Foryciarz, Moradshahi, and
  Lam}]{DBLP:conf/acl/CampagnaFML20}
Giovanni Campagna, Agata Foryciarz, Mehrad Moradshahi, and Monica~S. Lam. 2020.
\newblock \href {https://doi.org/10.18653/v1/2020.acl-main.12} {Zero-shot
  transfer learning with synthesized data for multi-domain dialogue state
  tracking}.
\newblock In \emph{Proceedings of the 58th Annual Meeting of the Association
  for Computational Linguistics, {ACL} 2020, Online, July 5-10, 2020}, pages
  122--132. Association for Computational Linguistics.

\bibitem[{Devlin et~al.(2019)Devlin, Chang, Lee, and
  Toutanova}]{DBLP:conf/naacl/DevlinCLT19}
Jacob Devlin, Ming{-}Wei Chang, Kenton Lee, and Kristina Toutanova. 2019.
\newblock \href {https://doi.org/10.18653/v1/n19-1423} {{BERT:} pre-training of
  deep bidirectional transformers for language understanding}.
\newblock In \emph{Proceedings of the 2019 Conference of the North American
  Chapter of the Association for Computational Linguistics: Human Language
  Technologies, {NAACL-HLT} 2019, Minneapolis, MN, USA, June 2-7, 2019, Volume
  1 (Long and Short Papers)}, pages 4171--4186. Association for Computational
  Linguistics.

\bibitem[{Einolghozati et~al.(2019)Einolghozati, Gupta, Mohit, and
  Shah}]{DBLP:journals/corr/abs-1911-05153}
Arash Einolghozati, Sonal Gupta, Mrinal Mohit, and Rushin Shah. 2019.
\newblock \href {http://arxiv.org/abs/1911.05153} {Improving robustness of task
  oriented dialog systems}.
\newblock \emph{CoRR}, abs/1911.05153.

\bibitem[{Gao et~al.(2020)Gao, Zhang, Ou, and Yu}]{DBLP:conf/acl/GaoZOY20}
Silin Gao, Yichi Zhang, Zhijian Ou, and Zhou Yu. 2020.
\newblock \href {https://doi.org/10.18653/v1/2020.acl-main.60} {Paraphrase
  augmented task-oriented dialog generation}.
\newblock In \emph{Proceedings of the 58th Annual Meeting of the Association
  for Computational Linguistics, {ACL} 2020, Online, July 5-10, 2020}, pages
  639--649. Association for Computational Linguistics.

\bibitem[{Gritta et~al.(2021)Gritta, Lampouras, and
  Iacobacci}]{DBLP:journals/tacl/GrittaLI21}
Milan Gritta, Gerasimos Lampouras, and Ignacio Iacobacci. 2021.
\newblock \href {https://doi.org/10.1162/tacl\_a\_00352} {Conversation graph:
  Data augmentation, training and evaluation for non-deterministic dialogue
  management}.
\newblock \emph{Trans. Assoc. Comput. Linguistics}, 9:36--52.

\bibitem[{Henderson et~al.(2014)Henderson, Thomson, and
  Young}]{DBLP:conf/sigdial/HendersonTY14}
Matthew Henderson, Blaise Thomson, and Steve~J. Young. 2014.
\newblock \href {https://doi.org/10.3115/v1/w14-4340} {Word-based dialog state
  tracking with recurrent neural networks}.
\newblock In \emph{Proceedings of the {SIGDIAL} 2014 Conference, The 15th
  Annual Meeting of the Special Interest Group on Discourse and Dialogue, 18-20
  June 2014, Philadelphia, PA, {USA}}, pages 292--299. The Association for
  Computer Linguistics.

\bibitem[{Hou et~al.(2018)Hou, Liu, Che, and Liu}]{DBLP:conf/coling/HouLCL18}
Yutai Hou, Yijia Liu, Wanxiang Che, and Ting Liu. 2018.
\newblock \href {https://aclanthology.org/C18-1105/} {Sequence-to-sequence data
  augmentation for dialogue language understanding}.
\newblock In \emph{Proceedings of the 27th International Conference on
  Computational Linguistics, {COLING} 2018, Santa Fe, New Mexico, USA, August
  20-26, 2018}, pages 1234--1245. Association for Computational Linguistics.

\bibitem[{Huang et~al.(2021)Huang, Li, Qu, and Pan}]{DBLP:conf/eacl/HuangLQP21}
Shuo Huang, Zhuang Li, Lizhen Qu, and Lei Pan. 2021.
\newblock \href {https://doi.org/10.18653/v1/2021.eacl-main.292} {On robustness
  of neural semantic parsers}.
\newblock In \emph{Proceedings of the 16th Conference of the European Chapter
  of the Association for Computational Linguistics: Main Volume, {EACL} 2021,
  Online, April 19 - 23, 2021}, pages 3333--3342. Association for Computational
  Linguistics.

\bibitem[{Kryscinski et~al.(2019)Kryscinski, Keskar, McCann, Xiong, and
  Socher}]{kryscinski-etal-2019-neural}
Wojciech Kryscinski, Nitish~Shirish Keskar, Bryan McCann, Caiming Xiong, and
  Richard Socher. 2019.
\newblock \href {https://doi.org/10.18653/v1/D19-1051} {Neural text
  summarization: A critical evaluation}.
\newblock In \emph{Proceedings of the 2019 Conference on Empirical Methods in
  Natural Language Processing and the 9th International Joint Conference on
  Natural Language Processing (EMNLP-IJCNLP)}, pages 540--551, Hong Kong,
  China. Association for Computational Linguistics.

\bibitem[{Lee et~al.(2021{\natexlab{a}})Lee, Cheng, and
  Ostendorf}]{DBLP:conf/emnlp/00010O21}
Chia{-}Hsuan Lee, Hao Cheng, and Mari Ostendorf. 2021{\natexlab{a}}.
\newblock \href {https://doi.org/10.18653/v1/2021.emnlp-main.404} {Dialogue
  state tracking with a language model using schema-driven prompting}.
\newblock In \emph{Proceedings of the 2021 Conference on Empirical Methods in
  Natural Language Processing, {EMNLP} 2021, Virtual Event / Punta Cana,
  Dominican Republic, 7-11 November, 2021}, pages 4937--4949. Association for
  Computational Linguistics.

\bibitem[{Lee et~al.(2021{\natexlab{b}})Lee, Gupta, Rastogi, Cao, Zhang, and
  Wu}]{DBLP:journals/corr/abs-2110-06800}
Harrison Lee, Raghav Gupta, Abhinav Rastogi, Yuan Cao, Bin Zhang, and Yonghui
  Wu. 2021{\natexlab{b}}.
\newblock \href {http://arxiv.org/abs/2110.06800} {{SGD-X:} {A} benchmark for
  robust generalization in schema-guided dialogue systems}.
\newblock \emph{CoRR}, abs/2110.06800.

\bibitem[{Lee et~al.(2022)Lee, Gupta, Rastogi, Cao, Zhang, and
  Wu}]{DBLP:conf/aaai/0001GR0ZW22}
Harrison Lee, Raghav Gupta, Abhinav Rastogi, Yuan Cao, Bin Zhang, and Yonghui
  Wu. 2022.
\newblock \href {https://ojs.aaai.org/index.php/AAAI/article/view/21341}
  {{SGD-X:} {A} benchmark for robust generalization in schema-guided dialogue
  systems}.
\newblock In \emph{Thirty-Sixth {AAAI} Conference on Artificial Intelligence,
  {AAAI} 2022, Thirty-Fourth Conference on Innovative Applications of
  Artificial Intelligence, {IAAI} 2022, The Twelveth Symposium on Educational
  Advances in Artificial Intelligence, {EAAI} 2022 Virtual Event, February 22 -
  March 1, 2022}, pages 10938--10946. {AAAI} Press.

\bibitem[{Lewis et~al.(2020)Lewis, Liu, Goyal, Ghazvininejad, Mohamed, Levy,
  Stoyanov, and Zettlemoyer}]{DBLP:conf/acl/LewisLGGMLSZ20}
Mike Lewis, Yinhan Liu, Naman Goyal, Marjan Ghazvininejad, Abdelrahman Mohamed,
  Omer Levy, Veselin Stoyanov, and Luke Zettlemoyer. 2020.
\newblock \href {https://doi.org/10.18653/v1/2020.acl-main.703} {{BART:}
  denoising sequence-to-sequence pre-training for natural language generation,
  translation, and comprehension}.
\newblock In \emph{Proceedings of the 58th Annual Meeting of the Association
  for Computational Linguistics, {ACL} 2020, Online, July 5-10, 2020}, pages
  7871--7880. Association for Computational Linguistics.

\bibitem[{Liu et~al.(2021)Liu, Takanobu, Wen, Wan, Li, Nie, Li, Peng, and
  Huang}]{DBLP:conf/acl/LiuTWWLNLPH20}
Jiexi Liu, Ryuichi Takanobu, Jiaxin Wen, Dazhen Wan, Hongguang Li, Weiran Nie,
  Cheng Li, Wei Peng, and Minlie Huang. 2021.
\newblock \href {https://doi.org/10.18653/v1/2021.acl-long.192} {Robustness
  testing of language understanding in task-oriented dialog}.
\newblock In \emph{Proceedings of the 59th Annual Meeting of the Association
  for Computational Linguistics and the 11th International Joint Conference on
  Natural Language Processing, {ACL/IJCNLP} 2021, (Volume 1: Long Papers),
  Virtual Event, August 1-6, 2021}, pages 2467--2480. Association for
  Computational Linguistics.

\bibitem[{Louvan and Magnini(2020)}]{DBLP:conf/paclic/LouvanM20}
Samuel Louvan and Bernardo Magnini. 2020.
\newblock \href {https://aclanthology.org/2020.paclic-1.20/} {Simple is better!
  lightweight data augmentation for low resource slot filling and intent
  classification}.
\newblock In \emph{Proceedings of the 34th Pacific Asia Conference on Language,
  Information and Computation, {PACLIC} 2020, Hanoi, Vietnam, October 24-26,
  2020}, pages 167--177. Association for Computational Linguistics.

\bibitem[{Ma et~al.(2019)Ma, Zeng, Zhu, Li, Yang, Yao, Zhou, and
  Shen}]{DBLP:journals/corr/abs-1912-09297}
Yue Ma, Zengfeng Zeng, Dawei Zhu, Xuan Li, Yiying Yang, Xiaoyuan Yao, Kaijie
  Zhou, and Jianping Shen. 2019.
\newblock \href {http://arxiv.org/abs/1912.09297} {An end-to-end dialogue state
  tracking system with machine reading comprehension and wide {\&} deep
  classification}.
\newblock \emph{CoRR}, abs/1912.09297.

\bibitem[{Maynez et~al.(2020)Maynez, Narayan, Bohnet, and
  McDonald}]{DBLP:conf/acl/MaynezNBM20}
Joshua Maynez, Shashi Narayan, Bernd Bohnet, and Ryan~T. McDonald. 2020.
\newblock \href {https://doi.org/10.18653/v1/2020.acl-main.173} {On
  faithfulness and factuality in abstractive summarization}.
\newblock In \emph{Proceedings of the 58th Annual Meeting of the Association
  for Computational Linguistics, {ACL} 2020, Online, July 5-10, 2020}, pages
  1906--1919. Association for Computational Linguistics.

\bibitem[{Mohapatra et~al.(2021)Mohapatra, Pandey, Contractor, and
  Joshi}]{DBLP:conf/emnlp/Mohapatra0CJ21}
Biswesh Mohapatra, Gaurav Pandey, Danish Contractor, and Sachindra Joshi. 2021.
\newblock \href {https://doi.org/10.18653/v1/2021.findings-emnlp.103}
  {Simulated chats for building dialog systems: Learning to generate
  conversations from instructions}.
\newblock In \emph{Findings of the Association for Computational Linguistics:
  {EMNLP} 2021, Virtual Event / Punta Cana, Dominican Republic, 16-20 November,
  2021}, pages 1190--1203. Association for Computational Linguistics.

\bibitem[{Narayan et~al.(2022)Narayan, Sim{\~{o}}es, Zhao, Maynez, Das,
  Collins, and Lapata}]{DBLP:conf/acl/NarayanSZM00L22}
Shashi Narayan, Gon{\c{c}}alo Sim{\~{o}}es, Yao Zhao, Joshua Maynez, Dipanjan
  Das, Michael Collins, and Mirella Lapata. 2022.
\newblock \href {https://aclanthology.org/2022.acl-long.94} {A well-composed
  text is half done! composition sampling for diverse conditional generation}.
\newblock In \emph{Proceedings of the 60th Annual Meeting of the Association
  for Computational Linguistics (Volume 1: Long Papers), {ACL} 2022, Dublin,
  Ireland, May 22-27, 2022}, pages 1319--1339. Association for Computational
  Linguistics.

\bibitem[{Quan and Xiong(2019)}]{DBLP:conf/ialp/QuanX19}
Jun Quan and Deyi Xiong. 2019.
\newblock \href {https://doi.org/10.1109/IALP48816.2019.9037690} {Effective
  data augmentation approaches to end-to-end task-oriented dialogue}.
\newblock In \emph{International Conference on Asian Language Processing,
  {IALP} 2019, Shanghai, China, November 15-17, 2019}, pages 47--52. {IEEE}.

\bibitem[{Raffel et~al.(2020)Raffel, Shazeer, Roberts, Lee, Narang, Matena,
  Zhou, Li, and Liu}]{DBLP:journals/jmlr/RaffelSRLNMZLL20}
Colin Raffel, Noam Shazeer, Adam Roberts, Katherine Lee, Sharan Narang, Michael
  Matena, Yanqi Zhou, Wei Li, and Peter~J. Liu. 2020.
\newblock \href {http://jmlr.org/papers/v21/20-074.html} {Exploring the limits
  of transfer learning with a unified text-to-text transformer}.
\newblock \emph{J. Mach. Learn. Res.}, 21:140:1--140:67.

\bibitem[{Rastogi et~al.(2020)Rastogi, Zang, Sunkara, Gupta, and
  Khaitan}]{DBLP:conf/aaai/RastogiZSGK20}
Abhinav Rastogi, Xiaoxue Zang, Srinivas Sunkara, Raghav Gupta, and Pranav
  Khaitan. 2020.
\newblock \href {https://ojs.aaai.org/index.php/AAAI/article/view/6394}
  {Towards scalable multi-domain conversational agents: The schema-guided
  dialogue dataset}.
\newblock In \emph{The Thirty-Fourth {AAAI} Conference on Artificial
  Intelligence, {AAAI} 2020, The Thirty-Second Innovative Applications of
  Artificial Intelligence Conference, {IAAI} 2020, The Tenth {AAAI} Symposium
  on Educational Advances in Artificial Intelligence, {EAAI} 2020, New York,
  NY, USA, February 7-12, 2020}, pages 8689--8696. {AAAI} Press.

\bibitem[{Sellam et~al.(2020)Sellam, Das, and
  Parikh}]{DBLP:conf/acl/SellamDP20}
Thibault Sellam, Dipanjan Das, and Ankur~P. Parikh. 2020.
\newblock \href {https://doi.org/10.18653/v1/2020.acl-main.704} {{BLEURT:}
  learning robust metrics for text generation}.
\newblock In \emph{Proceedings of the 58th Annual Meeting of the Association
  for Computational Linguistics, {ACL} 2020, Online, July 5-10, 2020}, pages
  7881--7892. Association for Computational Linguistics.

\bibitem[{Wei and Zou(2019)}]{DBLP:conf/emnlp/WeiZ19}
Jason~W. Wei and Kai Zou. 2019.
\newblock \href {https://doi.org/10.18653/v1/D19-1670} {{EDA:} easy data
  augmentation techniques for boosting performance on text classification
  tasks}.
\newblock In \emph{Proceedings of the 2019 Conference on Empirical Methods in
  Natural Language Processing and the 9th International Joint Conference on
  Natural Language Processing, {EMNLP-IJCNLP} 2019, Hong Kong, China, November
  3-7, 2019}, pages 6381--6387. Association for Computational Linguistics.

\bibitem[{Williams et~al.(2018)Williams, Nangia, and Bowman}]{N18-1101}
Adina Williams, Nikita Nangia, and Samuel Bowman. 2018.
\newblock \href {http://aclweb.org/anthology/N18-1101} {A broad-coverage
  challenge corpus for sentence understanding through inference}.
\newblock In \emph{Proceedings of the 2018 Conference of the North American
  Chapter of the Association for Computational Linguistics: Human Language
  Technologies, Volume 1 (Long Papers)}, pages 1112--1122. Association for
  Computational Linguistics.

\bibitem[{Wolf et~al.(2019)Wolf, Debut, Sanh, Chaumond, Delangue, Moi, Cistac,
  Rault, Louf, Funtowicz, and Brew}]{DBLP:journals/corr/abs-1910-03771}
Thomas Wolf, Lysandre Debut, Victor Sanh, Julien Chaumond, Clement Delangue,
  Anthony Moi, Pierric Cistac, Tim Rault, R{\'{e}}mi Louf, Morgan Funtowicz,
  and Jamie Brew. 2019.
\newblock \href {http://arxiv.org/abs/1910.03771} {Huggingface's transformers:
  State-of-the-art natural language processing}.
\newblock \emph{CoRR}, abs/1910.03771.

\bibitem[{Yin et~al.(2020)Yin, Shang, Jiang, Chen, and
  Liu}]{DBLP:conf/aaai/YinSJCL20}
Yichun Yin, Lifeng Shang, Xin Jiang, Xiao Chen, and Qun Liu. 2020.
\newblock \href {https://ojs.aaai.org/index.php/AAAI/article/view/6491} {Dialog
  state tracking with reinforced data augmentation}.
\newblock In \emph{The Thirty-Fourth {AAAI} Conference on Artificial
  Intelligence, {AAAI} 2020, The Thirty-Second Innovative Applications of
  Artificial Intelligence Conference, {IAAI} 2020, The Tenth {AAAI} Symposium
  on Educational Advances in Artificial Intelligence, {EAAI} 2020, New York,
  NY, USA, February 7-12, 2020}, pages 9474--9481. {AAAI} Press.

\bibitem[{Zhang et~al.(2020{\natexlab{a}})Zhang, Zhao, Saleh, and
  Liu}]{DBLP:conf/icml/ZhangZSL20}
Jingqing Zhang, Yao Zhao, Mohammad Saleh, and Peter~J. Liu. 2020{\natexlab{a}}.
\newblock \href {http://proceedings.mlr.press/v119/zhang20ae.html} {{PEGASUS:}
  pre-training with extracted gap-sentences for abstractive summarization}.
\newblock In \emph{Proceedings of the 37th International Conference on Machine
  Learning, {ICML} 2020, 13-18 July 2020, Virtual Event}, volume 119 of
  \emph{Proceedings of Machine Learning Research}, pages 11328--11339. {PMLR}.

\bibitem[{Zhang et~al.(2020{\natexlab{b}})Zhang, Ou, and
  Yu}]{DBLP:conf/aaai/ZhangOY20}
Yichi Zhang, Zhijian Ou, and Zhou Yu. 2020{\natexlab{b}}.
\newblock \href {https://ojs.aaai.org/index.php/AAAI/article/view/6507}
  {Task-oriented dialog systems that consider multiple appropriate responses
  under the same context}.
\newblock In \emph{The Thirty-Fourth {AAAI} Conference on Artificial
  Intelligence, {AAAI} 2020, The Thirty-Second Innovative Applications of
  Artificial Intelligence Conference, {IAAI} 2020, The Tenth {AAAI} Symposium
  on Educational Advances in Artificial Intelligence, {EAAI} 2020, New York,
  NY, USA, February 7-12, 2020}, pages 9604--9611. {AAAI} Press.

\bibitem[{Zhao et~al.(2022)Zhao, Gupta, Cao, Yu, Wang, Lee, Rastogi, Shafran,
  and Wu}]{DBLP:journals/corr/abs-2201-08904}
Jeffrey Zhao, Raghav Gupta, Yuan Cao, Dian Yu, Mingqiu Wang, Harrison Lee,
  Abhinav Rastogi, Izhak Shafran, and Yonghui Wu. 2022.
\newblock \href {http://arxiv.org/abs/2201.08904} {Description-driven
  task-oriented dialog modeling}.
\newblock \emph{CoRR}, abs/2201.08904.

\bibitem[{Zhu et~al.(2018)Zhu, Lu, Zheng, Guo, Zhang, Wang, and
  Yu}]{DBLP:conf/sigir/ZhuLZGZWY18}
Yaoming Zhu, Sidi Lu, Lei Zheng, Jiaxian Guo, Weinan Zhang, Jun Wang, and Yong
  Yu. 2018.
\newblock \href {https://doi.org/10.1145/3209978.3210080} {Texygen: {A}
  benchmarking platform for text generation models}.
\newblock In \emph{The 41st International {ACM} {SIGIR} Conference on Research
  {\&} Development in Information Retrieval, {SIGIR} 2018, Ann Arbor, MI, USA,
  July 08-12, 2018}, pages 1097--1100. {ACM}.

\end{thebibliography}
\bibliographystyle{acl_natbib}

\appendix

\section{The SGD and SGD-X datasets}
\label{appendix:sgd-sgd-x}
\textbf{SGD} As mentioned in Section \ref{sec:intro}, the conversations in the SGD dataset are grounded in schemata, which describe a set of service APIs. The most important schema \textit{elements} are\footnote{Examples below are taken from the SGD test set.}:
\begin{itemize}
    \item a \textit{service name} (e.g. \textit{Messaging\_1}) followed by a \textit{service description} (e.g. \textit{Connect and share locations with your contacts})
    \item one or more API functions to be invoked as users solve tasks, referred to as \textit{(user) intents}; each intent has a \textit{name} (e.g. \textit{ShareLocation}) and an \textit{intent description} (e.g. \textit{Send your location to a contact})
    \item optional and required arguments for each API function, or \textit{slots}; each slot has a \textit{name} (e.g. \textit{location}) and a \textit{slot description} (e.g. \textit{Location to share with the contact})
\end{itemize}

\textbf{SGD-X} \citet{DBLP:conf/aaai/0001GR0ZW22} observe that $71\%$ of intent names and $65\%$ of slot names from unseen APIs exactly match the train set. Furthermore, descriptions are stylistically uniform across the train and test sets. For example,  all boolean slots begin with the phrase \textit{Boolean flag ...} or \textit{Whether...}. Therefore, they create the SGD-X dataset as follows:
\begin{itemize}
    \item crowdsource schema element paraphrasing to more than $400$ authors via Amazon Mechanical Turk. Each crowdworker either paraphrases all names or all descriptions for a given schema
    \item manually vet responses for quality and correctness.
\end{itemize}
The slot names collected are sorted in increasing order of their Levenshtein distance to the SGD slot names whereas the descriptions are sorted according to the Jaccard distance between their lemmatized forms (excluding stop words).
An example of SGD-X description paraphrases in shown in Table \ref{tab:sgd-x-descr}.
\begin{table}[h]
\setlength{\tabcolsep}{2pt}
\centering
\small
\begin{tabular}{cc}
\toprule
Variant & Description                                               \\ \hline
SGD     & Category to which the attraction belongs                  \\
v1      & The category that describes what kind of attraction it is \\
v2      & Category of place of interest                             \\
v3      & Type of tourist attraction                                \\
v4      & Choose the kind of tourist landmark                       \\
v5      & The kind of tourist hotspot                 \\   \bottomrule          
\end{tabular}
\caption{Example of descriptions paraphrases from the SGD-X test schemas. The more similar \texttt{v1} description contains overlapping vocabulary with the SGD test set description, whereas \texttt{v4} and \texttt{v5} variants are dissimilar both stylistically and lexically }
\label{tab:sgd-x-descr}
\end{table}

While the examples above are paraphrases of the SGD input, in general, the semantic content of the schema element paraphrases is not perfectly overlapping with the input as the crowdworkers use information from the wider service context when creating new elements.

\begin{table*}[]
\centering
\small  
\begin{tabular}{@{}lc@{}}
\toprule
\multicolumn{1}{c}{Filter name}         & Filtered example                                         \\ \midrule
\texttt{contains advice}                & An appointment \textbf{is necessary for your hair}.      \\
\texttt{describes action}               & They commemorate the number of flights to the airport.   \\
\texttt{has named entities}             & Enter the doctor's \textbf{Leningrad} address.            \\
\texttt{has low frequency words}        & The address is \textbf{ofadvisory}.                       \\
\texttt{discard multiple sentences}   & \multicolumn{1}{l}{The address is the dentist's box. Guidelines for hiring a dentist.} \\
\texttt{has repeated ngrams}            & \textbf{The dentist} is Address of \textbf{the dentist}. \\
\texttt{has repeated similar bigrams} & The \textbf{type of event} is stated in the title \textbf{of the event}.                \\
\texttt{has consecutive repeated words} & Average review rating for a hotel hotel.                 \\
\texttt{is past tense sentence}         & It was the dentist's address.                             \\
\texttt{is passive voice sentence}      & The address was given by abrasives from the dentist.     \\
\texttt{is question}                    & Is there a balance of the account?                       \\
\texttt{has alphanumeric words}         & \textbf{400} baths in an apartment.                      \\ \bottomrule
\end{tabular}
\caption{Filters implemented along with sample examples they discard}
\label{tab:heuristic-filters}
\end{table*}

\section{Ranking Framework}
\label{appendix:ranking-framework}

\subsection{Candidate generation}
\label{appendix:cand-gen}
Algorithm \ref{algo:generate} summarises the candidate generation procedure, which takes any paraphrase model, a list of model-specific generation parameters and, optionally, a list of filters as an input (line 1). These parameters are temperature and number of beams for Pegasus, or a grid of lexical, semantic and syntactic distances for the Quality Controlled Paraphrase Generation (QCPG) \cite{DBLP:conf/acl/BandelASSSE22} model presented in Appendix \ref{appdx:QCPG-BLEURT-RESULTS}. The model generates one or more paraphrases, which are filtered before returning (lines 3-8). We describe the filtering process next.

\textbf{Heuristic filtering}
Our main motivation for implementing heuristic filters is to filter the majority of poor quality candidates, without making use of the large GPU cards required to run the entailment model. We also address the fact that the model is free to generate a very large number of candidates and therefore is expected to hallucinate significantly. These filters are general purpose and are implemented in few lines of code using the \texttt{spaCy} and \texttt{nltk} libraries. Table  \ref{tab:heuristic-filters} lists active filters along with typical examples filtered.

\textbf{Entailment filtering} We implement our entailment filter using BART \cite{DBLP:conf/acl/LewisLGGMLSZ20}\footnote{Avaialble at \url{https://huggingface.co/facebook/bart-large-mnli}.}. This model is pre-trained on the MNLI dataset \cite{N18-1101}. To measure entailment this model consumes a premise and hypothesis in the format \texttt{premise <SEP> hypothesis}. In our implementation we replace \texttt{premise} with the description to be paraphrased. By default, the \texttt{hypothesis} is a template of the form \texttt{This example is \{\}.}, where \texttt{\{\}} is a placeholder for the user hypothesis, in our case the paraphrased description. We find that considering alternative templates improves the reliability of the model, so we consider \texttt{\{\}}, \texttt{This example has the same meaning as \texttt \{\}.}, \texttt{This text is about \{\}.}, and \texttt{This example implies that \{\}.}, averaging the entailment scores across templates to calculate the entailment score. The same procedure is followed when computing the entailment of candidates during ranking.

\begin{algorithm}
\SetNlSty{}{}{:}
\SetAlgoNlRelativeSize{-1}
\AlgoDontDisplayBlockMarkers\SetAlgoNoEnd\SetAlgoNoLine%
\caption{Candidate generation}\label{algo:generate}
\Fn{\generate{\texttt{model}: Any, \texttt{inp}: str, \texttt{params}: dict, \texttt{filters}: Optional[list[Callable]]}}{
\KwData{\texttt{model} text generation model, \texttt{inp} input sentence, \texttt{params} model specific parameters, \texttt{filters} a list of boolean functions}
\KwResult{\texttt{cands} list of \texttt{inp} paraphrases}

\texttt{cands} $\gets$ [] \;
\For{\texttt{p} \KwTo \texttt{params}}{
    c $\gets$ \texttt{model.forward(inp, **p)} \;
     c $\gets$ [p \textbf{for} p \KwTo c \textbf{if} \KwNot \Any{\texttt{f(p, inp) \textbf{for} \texttt{f} \KwTo \texttt{filters}}}]

    \texttt{cands.extend(c)} 
}
\Return \texttt{cands}
}
\end{algorithm}

\subsection{Ranking}

\begin{algorithm}
\small
\SetNlSty{}{}{:}
\SetAlgoNlRelativeSize{-1}
\AlgoDontDisplayBlockMarkers\SetAlgoNoEnd\SetAlgoNoLine%
\caption{Tree ranking}
\label{algo:ranking}
\Fn{\rank{\texttt{root}: Node, \texttt{n}: int, \texttt{f\_dec}: list[Callable]}}{
\KwData{\texttt{root}, \texttt{n} number of candidates, \texttt{f\_dec} decision functions}
\KwResult{list of \texttt{n} ranked candidates}

\texttt{ranked} $\gets$ \syntax(\texttt{root}) \;
\texttt{n} $\gets$  \texttt{n} - \texttt{len(\texttt{ranked})} \;

\While{\texttt{len}(\texttt{ranked}) $\neq$ \texttt{n}}{
\For{\texttt{next} \KwTo \bfs(\texttt{root})}{
    \For{\texttt{f} \KwTo \texttt{f\_dec}}{
      
    }
    \texttt{cand} $\gets$ \select(\texttt{next.sents}) \;
    \texttt{ranked.add(cand)} \;
    \prune (\texttt{next}, \texttt{cand}) \;
}
}
\Return \texttt{ranked}
}
\end{algorithm}

\textbf{Ranking} Algorithm \ref{algo:ranking} summarises the tree-ranking procedure. This procedure takes as an input the tree constructed as described in Algorithm \ref{algo:build}, along with a list of decision functions \texttt{f\_dec}. Our algorithm starts by selecting a paraphrase via depth first traversal of the subtree rooted at $J=0$ (line 2). The remainder of the candidates are selected by traversing the first level in a breadth-first manner (line 5) and depth-first traversal of each subtree returned during the level-order traversal (lines 6-8). Here the semantics of \texttt{f(next.children)} is that the decision function $\texttt{f}$ takes all the children of \texttt{next} as input and returns a single node which is next in the traversal. A candidate is selected from the leaf\footnote{There can be multiple, possibly repeated candidates in a leaf because the generative model may generate the same output given different parameter settings. We select the most common one if there are repeated candidates and randomly otherwise.}(line 8) and subsequently removed from the candidates list (line 10). This is to avoid selecting the same candidate multiple times in situations where the paraphrase model generates few distinct candidates.
\begin{figure*}[t]
\centering
\begin{subfigure}[]{\textwidth}
    \centering
    \includegraphics[width=\textwidth]{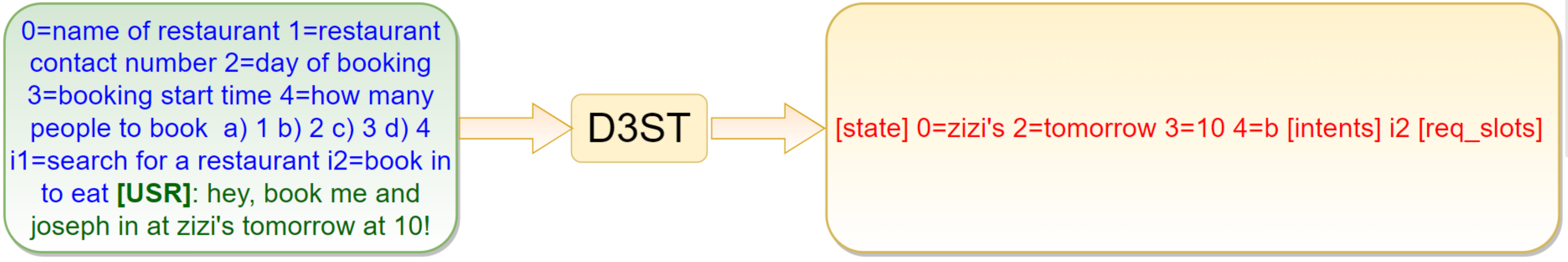}
    \caption{Visual representation of D3ST inputs and targets. Blue font \textcolor{blue}{blue}, preceding the \texttt{[USR] special token represents the prompt, consisting of slot descriptions extracted from the schema. Each slot is assigned an index, which is used to recover the slot value pairs during post-processing. Note that slot \texttt{4} is categorical, and so the string \texttt{a) 1 b) 2 c) 3 d) 4}} is appended to the description to indicate the model that it should output one of the choices. The dialogue history follows the \texttt{[USR]} token. The \textcolor{red}{model outputs} only active slots, in this case omitting slot $1$ because is has not been mentioned. The entire dialogue state (as opposed to turn state) is generated at every turn.}
    \label{fig:d3st}
\end{subfigure}
\smallskip 
\begin{subfigure}[]{\textwidth}
    \centering
    \includegraphics[width=\textwidth]{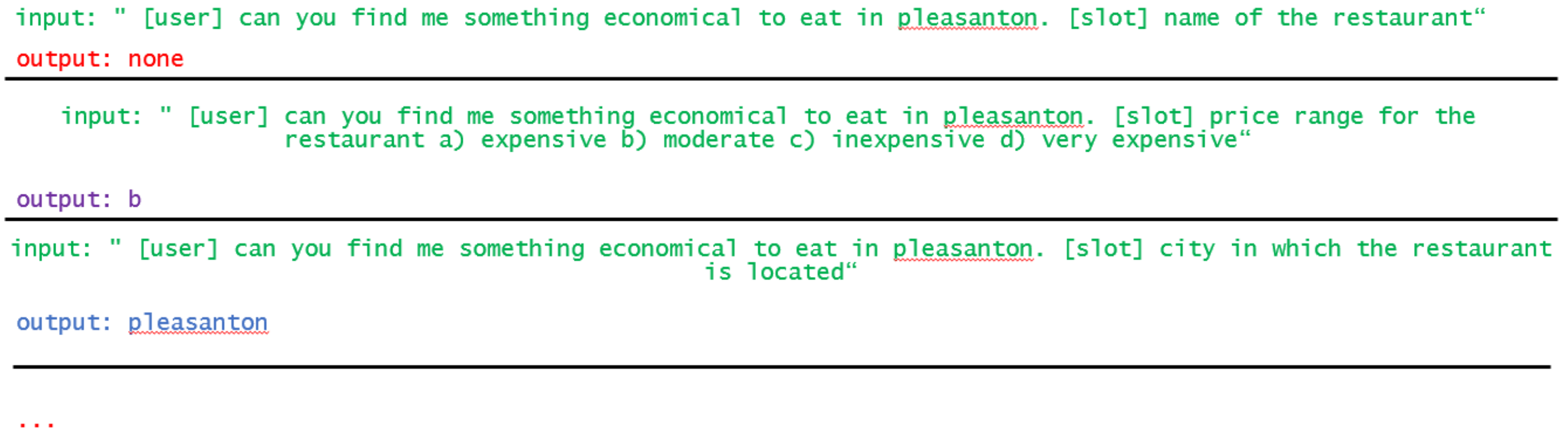}
    \caption{Visual representation of T5DST inputs and targets. The string \texttt{[slot]} separates the dialogue history from the slot description. In the first example the model outputs the special value \textcolor{red}{\texttt{none}} to indicate that the slot was not mentioned by the user. For categorical slots (second row from the top), the slot description is concatenated with options containing all possible slot values and the model predicts the correct option. For non-categorical slots (third row), the exact value is predicted. The ellipsis indicates that  \textcolor{red}{\texttt{none}} is predicted for all other slots in the \texttt{Restaurant\_1} schema that are not mentioned in the dialogue history}
    \label{fig:t5dst}
\end{subfigure}
\caption{Prompt formats for a)  D3ST  b) T5DST }
\label{fig:prompting}
\end{figure*}

\section{State Tracking Baselines}
\label{appendix:state-tracking}
\textbf{D3ST} We process the data as described by \citet{DBLP:journals/corr/abs-2201-08904} with the following differences:

\begin{itemize}
    \item The indices are separated by the \texttt{=} symbol in both the inputs and the targets, to avoid a parsing ambiguity which occurs for time slots if \texttt{:} is used as a separator for targets
    \item For categorical slots which take the \texttt{dontcare} special value, our output contains \texttt{slot\_index: dontcare} substring and we do not include the \texttt{dontcare} value in the prefix together with the other options
    \item We lowercase the inputs and the targets\footnote{This appears in illustrations but is not explicitly stated by \citep{DBLP:journals/corr/abs-2201-08904}.}.
\end{itemize}

We obtain $175,780$ examples from the original SGD dataset, which are truncated to the last $1,024$ tokens on the input side. See Figure \ref{fig:d3st} for a visual representation of the model inputs and outputs. We optimise the model using the Adafactor optimizer and effective batch size $32$, starting from the initial weights \texttt{google/t5-v1\_1-base} published by \texttt{huggingface} \cite{DBLP:journals/corr/abs-1910-03771}. We interpolate the learning rate linearly between $0$ and $10^{-4}$ over the first $1000$ steps and keep it constant thereafter. We select the model by evaluating the development set JGA every $5000$ gradient updates, stopping the training if said metric fails to improve after $3$ consecutive evaluations. All numbers in Table \ref{tab:dst-results} are averages of 3 runs, except the SGD-X experiment for T5DST which is a single run.

\textbf{T5DST} Given a dialogue in the SGD training set we consider all partial dialogue histories $\{u_1, s_1, ... s_{t-1}, u_t\}$ with $t \in \overline{0, T}$ where $T$ is maximum index of the user turn in a dialogue. The turns in each dialogue history are lowercased and separated \texttt{[usr]} and \texttt{[sys]} tokens, \textit{not treated} as special tokens. For each dialogue history we create a training example for each slot in the ground truth schema, which contains the concatenated turns suffixed with the string \texttt{[slot] [slot\_description]} where the placeholder \texttt{[slot\_description]} is replaced by the lowercase descriptions extracted from the SGD schemata. This yields $1,601,356$ examples for the SGD training dataset. See Figure \ref{fig:t5dst} for a representation of the model inputs and outputs.

We optimise the model using the Adafactor optimizer and effective batch size $256$, starting from the initial weights \texttt{google/t5-v1\_1-base} published by \texttt{huggingface}. We interpolate the learning rate linearly between $0$ and $10^{-4}$ over the first $1000$ steps and keep it constant thereafter. We perform $20,000$ optimisation steps\footnote{For a single run, this is approximately $6$ hours of computation on $8$ \texttt{nvidia A100-80GB} cards. Moreover, decoding a single run on SGD and SGD-X takes $6$ hours.}, limiting the number of training steps to $40,000$ steps\footnote{This is sufficient so that the model sees every example once when working with an augmented training set six times the size of SGD.} for all augmented data experiments. All numbers in Table \ref{tab:dst-results} are averages of 3 runs, except the Oracle experiment on T5DST which is a single run.
\begin{table*}[h!]
\centering
\begin{minipage}[b]{0.98\columnwidth}
    \centering
    \small
\resizebox{\columnwidth}{!}{%
    \begin{tabular}{cclllcc}
\bottomrule
Metric &
  Ranking &
  \multicolumn{1}{c}{v1} &
  \multicolumn{1}{c}{v2} &
  \multicolumn{1}{c}{v3} &
  v4 &
  v5 \\ \hline
\multirow{2}{*}{Jaccard Dist} &
  Backtr. 4x &
  18.2 &
  29.9 &
  43.9 &
  - &
  - \\
 &
  Backtr. 6x &
  12.9 &
  22.7 &
  27.8 &
  \multicolumn{1}{l}{35.6} &
  \multicolumn{1}{l}{46.7} \\ \hline
\multirow{2}{*}{Entailment} &
  Backtr. 4x &
  97.5 &
  96.5 &
  95.9 &
  - &
  - \\
 &
  Backtr. 6x &
  98.0 &
  97.5 &
  95.2 &
  \multicolumn{1}{l}{94.8} &
  \multicolumn{1}{l}{95.5} \\ \hline
\multirow{2}{*}{BLEU} &
  Backtr. 4x &
  36.4 &
  26.01 &
  18.9 &
  - &
  - \\
 &
  Backtr. 6x &
  51.3 &
  37.2 &
  29.5 &
  \multicolumn{1}{l}{23.4} &
  \multicolumn{1}{l}{18.2} \\ \hline
\multirow{2}{*}{self-BLEU} & Backtr. 4x & \multicolumn{1}{c}{-} & \multicolumn{1}{c}{49.3} & \multicolumn{1}{c}{41.7} & -    & -    \\
                           & Backtr. 6x & \multicolumn{1}{c}{-} & \multicolumn{1}{c}{55.3} & \multicolumn{1}{c}{49.7} & 44.6 & 39.6 \\ \bottomrule

\end{tabular}
}
\caption{Effect of using French and Russian as additional pivot languages on automatic metrics}
\label{tab:backtr-4x-6x}
\end{minipage}\hfill
\begin{minipage}[b]{0.98\columnwidth}
\flushleft
\small
\resizebox{\columnwidth}{!}{
\begin{tabular}{ccllllc}

\toprule
Metric & Ranking & \multicolumn{1}{c}{v1} & \multicolumn{1}{c}{v2} & \multicolumn{1}{c}{v3} & \multicolumn{1}{c}{v4} & v5 \\ \hline
\multirow{3}{*}{Jaccard Dist}                   & Pegasus + BART   & 13.0 & 61.2 & 68.8  & 71.2 & \multicolumn{1}{l}{76.2} \\
                                                & Pegasus B + BART & 10.2 & 38.3 & 46.9  & 55.1 & \multicolumn{1}{l}{54.5} \\
                                                & SGD-X            & 55.6 & 65.6 & 71.2  & 78.1 & \multicolumn{1}{l}{85.7} \\ \hline
\multicolumn{1}{l}{\multirow{3}{*}{Entailment}} & Pegasus + BART   & 99.1 & 96.3 & 94.6  & 94.2 & 94.2                     \\
\multicolumn{1}{l}{}                            & Pegasus B + BART & 98.8 & 98.2 & 96..4 & 96.2 & 96.7                     \\
\multicolumn{1}{l}{}                            & SGD-X            & 89.7 & 88.0 & 88.4  & 86.8 & 87.5                     \\ \bottomrule
\end{tabular}
}
\caption{Comparison of diversity and semantic faithfulness metrics for slot description paraphrases}
\label{tab:ranking-algos-comparison-slots}
\end{minipage}
\end{table*}

\section{Additional Results}
\subsection{Increasing backtranslation dataset size}
\label{appendix:additional-results}
\begin{table*}[!h]
\centering
\small
\begin{tabular}{ccccccc}
\toprule
Index &
  Augmentation &
  $\mbox{JGA}_{orig}$ &
  $\mbox{JGA}_{v_{1-5}}$ &
  $\mbox{JGA}_{v_{1-5}}^{seen}$ &
  $\mbox{JGA}_{v_{1-5}}^{unseen}$ &
  $\mbox{SS}_{JGA}$ \\ \hline
1 &
  Pegasus+BART 6x &
  71.2 &
  \underline{63.9} &
  \underline{85.9} &
  {\textbf{56.6} } &
  { \textbf{46.5}} \\ 
2 &
  Pegasus+BLEURT 6x &
  \underline{72.4} &
  {\textbf{64.0}} &
  {\textbf{86.6} } &
  {\color[HTML]{000000} \underline {56.4}} &
  {\color[HTML]{000000} \underline{46.6}} \\ 
3 &
  QCPG+BLEURT 6x &
  { \textbf{72.7}} &
  {\color[HTML]{000000} 63.2} &
  {\color[HTML]{000000} 85.2} &
  {\color[HTML]{000000} 55.9} &
  {\color[HTML]{000000} 47.3} \\ \hline
4 &
  Backtranslation 4x \citep{DBLP:journals/corr/abs-2110-06800} &
  72.1 &
  62.2 &
  84.0 &
  54.9 &
  53.1 \\ \bottomrule
\end{tabular}
\caption{Ranking with a more accurate semantic faithfulness metric (row 2) or generating candidates with a controllable paraphrase model (row 4) can be used to boost SGD performance over our Pegasus+BART approach (row 1). Bold font marks column maximum, underlined second largest number.}
\label{tab:qcpg-bleurt}
\end{table*}
We include Table \ref{tab:backtr-4x-6x} to substantiate our intuition that the training with the Backtranslation 6x scheme does not yield further improvement compared to the Backtranslation 4x scheme as the additional data does not significantly increase the prompt diversity. Most clearly, this is indicated by the fact that the $v5$ variant has similar BLEU to variant $v3$ in Backtranslation 4x, indicating that a large proportion of additional data has some overlaps more with the SGD distribution than the data backtranslated to Chinese, Korean and Japanese. This is also indicated by how self-BLEU decays as more data is added, comparatively, between \texttt{Backtranslation 4x} and \texttt{Backtranslation 6x}. 
\subsection{Controlling schema generation diversity}
\label{appdx:control}
Table \ref{tab:ranking-algos-comparison-slots} shows that the alternative schema scheme generates schemas with lower average Jaccard distance and higher entailment with respect to the SGD schemata. We find this effectively controls the noise in the data, leading to improved performance for T5DST and similar performance to \texttt{PEGASUS + BART} for D3ST.

\section{Schema Generation with BLEURT and QCPG}
\label{appdx:QCPG-BLEURT-RESULTS}

BLEURT \citep{DBLP:conf/acl/SellamDP20} is a BERT-based natural metric commonly used in translation, so it is expected to be highly sensitive to semantic differences. In Table \ref{tab:qcpg-bleurt} we show that simply re-ranking the Pegasus output space with BLEURT improves SGD performance comparably with backtranslation (rows 2\&4) and the robustness and generalisation improvements are maintained. 

\citet{DBLP:conf/acl/BandelASSSE22} exploit high quality examples in paraphrase corpora by conditioning the model with a string \textit{quality parameters string} outlining target semantic, syntactic and lexical distances of the generated paraphrase during finetuning. At inference one must specify these parameters to obtain diverse yet high quality paraphrases. We could not apply the quality parameter selection method proposed by QCPG authors at inference time as the code had not been fully released at the time of writing. Instead, we generated a large number of paraphrases with different quality targets and greedy decoding, and re-ranked the candidates using our framework. This demonstrates the versatility of our framework. In Table \ref{tab:qcpg-bleurt} we show that this model can equally achieve improved performance on SGD. The improvement on SGD-X is slightly less than achieved by \texttt{PEGASUS+BART 6x}, as expected since greedy decoding and better semantic faithfulness optimisation generate schemata closer to the SGD distribution so less out-of-distribution improvement is achieved.

This experiments in this section and Appendix \ref{appdx:control} demonstrate the versatility of our framework and its usefulness as a tool for generating synthetic schema prompts.

\end{document}